\documentclass[letterpaper]{article} 
\usepackage{aaai2026}  
\usepackage{times}  
\usepackage{helvet}  
\usepackage{courier}  
\usepackage[hyphens]{url}  
\usepackage{graphicx} 
\usepackage{booktabs}
\usepackage{multirow}
\usepackage{makecell}
\usepackage{amsmath}
\usepackage{amsfonts}
\usepackage{tabularx}
\usepackage{array} 
\usepackage{natbib}  
\usepackage{caption} 
\usepackage{algorithm}
\usepackage{algorithmic}

\usepackage{newfloat}
\usepackage{listings}
\DeclareCaptionStyle{ruled}{labelfont=normalfont,labelsep=colon,strut=off} 
\floatstyle{ruled}
\newfloat{listing}{tb}{lst}{}
\floatname{listing}{Listing}
\title{BrainHGT: A Hierarchical Graph Transformer for Interpretable \\ Brain Network Analysis}
\author{
    Jiajun Ma,
    Yongchao Zhang,
    Chao Zhang,
    Zhao Lv,
    Shengbing Pei\thanks{Corresponding author}\\
}
\affiliations{
    Anhui Province Key Laboratory of Multimodal Cognitive Computation, \\School of Computer Science and Technology, Anhui University\\
    
e23301163@stu.ahu.edu.cn,
e24301188@stu.ahu.edu.cn,
iiphci\_ahu@163.com,
kjlz@ahu.edu.cn,
shengbingpei@ahu.edu.cn

}

\begin{document}

\maketitle

\begin{abstract}

Graph Transformer shows remarkable potential in brain network analysis due to its ability to model graph structures and complex node relationships. Most existing methods typically model the brain as a flat network, ignoring its modular structure, and their attention mechanisms treat all brain region connections equally, ignoring distance-related node connection patterns. However, brain information processing is a hierarchical process that involves local and long-range interactions between brain regions, interactions between regions and sub-functional modules, and interactions among functional modules themselves. This hierarchical interaction mechanism enables the brain to efficiently integrate local computations and global information flow, supporting the execution of complex cognitive functions. To address this issue, we propose BrainHGT, a hierarchical Graph Transformer that simulates the brain’s natural information processing from local regions to global communities. Specifically, we design a novel long-short range attention encoder that utilizes parallel pathways to handle dense local interactions and sparse long-range connections, thereby effectively alleviating the over-globalizing issue. To further capture the brain’s modular architecture, we designe a prior-guided clustering module that utilizes a cross-attention mechanism to group brain regions into functional communities and leverage neuroanatomical prior to guide the clustering process, thereby improving the biological plausibility and interpretability. Experimental results indicate that our proposed method significantly improves performance of disease identification, and can reliably capture the sub-functional modules of the brain, demonstrating its interpretability.

\end{abstract}

\begin{links}
    \link{Code}{https://github.com/null-cks/BrainHGT}
\end{links}

\section{Introduction}
\begin{figure}[t]
\centering
\includegraphics[width=1\columnwidth]{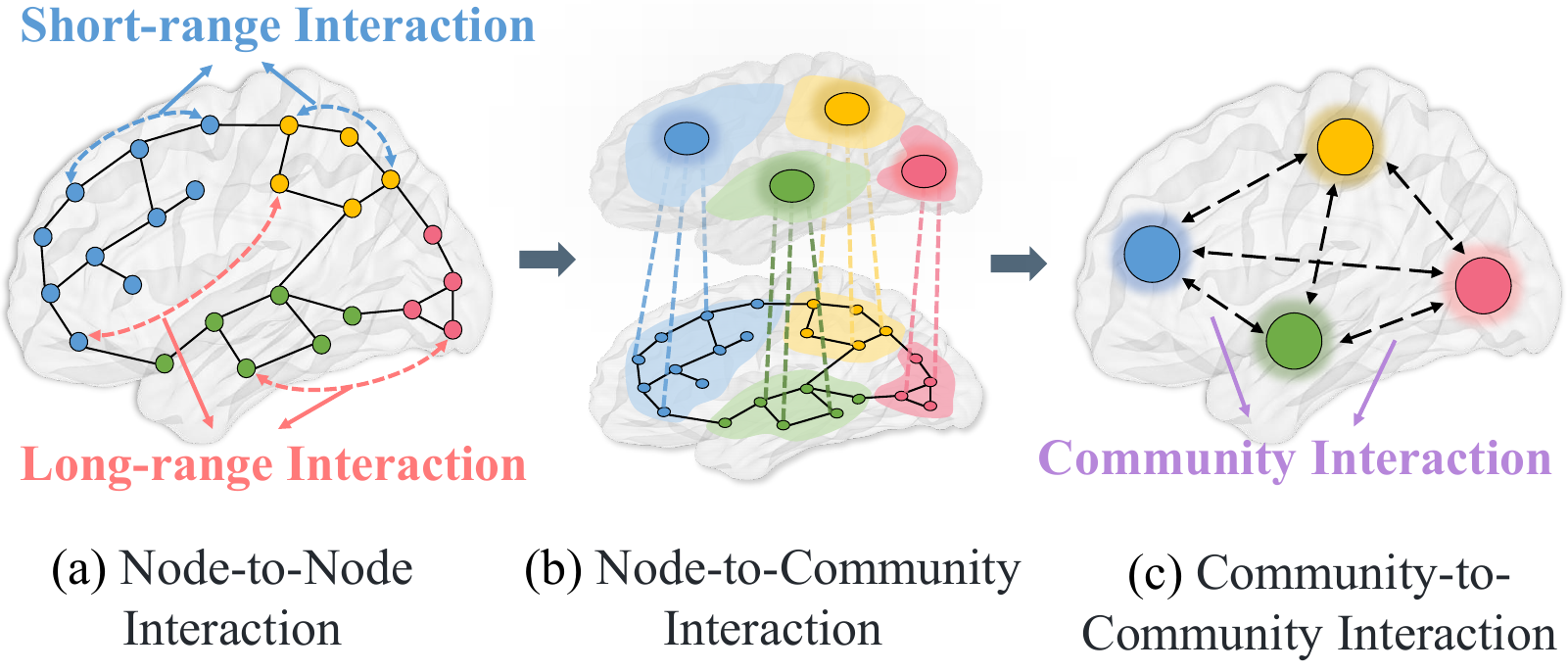}
\caption{The hierarchical model of brain information processing, showing three levels of interaction.}
\label{fig:Hierarchical}
\end{figure}
The diagnosis of neurological and psychiatric disorders, such as autism spectrum disorder (ASD) and Alzheimer's disease (AD), faces a significant challenge due to the lack of objective and reliable biomarkers \cite{lord2020autism, zhang2024review}. Resting-state functional magnetic resonance imaging (fMRI) has emerged as an important neuroimaging tool to address this problem. The non-invasive technique measures blood-oxygen-level-dependent (BOLD) signals to construct the brain's functional connectivity network, providing a unique insight into the brain's intrinsic organization and, more critically, helping to identify biomarkers that are crucial for the diagnosis and characterization of these diseases \cite{li2019graph,xu2024contrastive, Kong2024msstan}.

For the brain network analysis, graph neural networks (GNNs) are a commonly used class of methods \cite{zhang2024modeling,luo2024graph}. By modeling brain regions of interest (ROIs) as nodes and their connections as edges, GNNs automatically learn discriminative features directly from complex brain network topologies, bypassing extensive and often subjective manual feature engineering \cite{zhang2022classification}. Despite this advantage, GNNs are inherently limited by a receptive field confined to a local $k$-hop neighborhood, this make them ineffective at capturing the long-range dependencies that are crucial to the brain's large-scale functional organization. In contrast, Graph Transformer \cite{Cai2023Graph,song2024knowledge,Qu2024GGT} can model interactions between any pair of ROIs via self-attention mechanism, naturally capturing the global long-range connectivity patterns inherent in brain networks.

However, the global self-attention of Graph Transformer brings new challenges. Recent studies reveal that standard Graph Transformer exhibits a tendency towards over-globalizing, where attention is indiscriminately distributed across all nodes \cite{xing2024less}. This mechanism is inconsistent with the small-world network properties of the brain, where connectivity is heavily biased towards local, dense connections. And the probability of connection decays exponentially with distance, global communication is supplemented only by sparse yet critical long-range pathways \cite{bassett2017small,bullmore2012economy}. Consequently, the over-globalizing effect in Graph Transformer can dilute functionally relevant signals from nearby neighbors with irrelevant information from distant nodes, weakening the model's representational capacity. To address this, we process local and global information in parallel, using a learnable decay mask to strengthen local interactions, this effectively mitigates the over-globalizing problem and produces richer ROI-level spatial representations.

A more critical issue is that prevailing methods treat the brain as a flat, non-hierarchical graph structure, neglecting its inherent modular properties. In reality, the functional organization of brain is not confined to singular inter-regional interactions, but relies on higher-level coordination between functional modules to achieve whole-brain functions \cite{sporns2016modular,betzel2017multi}. The brain's information processing follows a natural hierarchical pattern. As shown in Figure \ref{fig:Hierarchical}, it begins with local and long-range interactions between brain regions, then organizes into large-scale networks (communities) with distinct functions, and finally achieves complex cognitive and behavioral functions through the dynamic coordination of these communities. Due to the lack of effective modeling for this hierarchical process, the decision-making process of current methods is often not comprehensive enough. Furthermore, the learned features typically lack clear neuroscientific meaning, severely hinders the method's potential for clinical translation and scientific discovery. 
To capture this hierarchical structure, we propose a prior-guided clustering module that uses neuroanatomical knowledge to guide the integration of information from microscopic nodes to macroscopic communities, and subsequently models community-level interactions to improve the biological plausibility and interpretability of the learning process.

Here, we propose BrainHGT, a novel Graph Transformer framework that simulates hierarchical brain processing to enhance interpretability. The main contributions include:

\begin{itemize}

\item A long-short range attention (LSRA) module is designed to balance strengthening functionally relevant local interactions and capturing sparse critical global dependencies, thereby generating richer node representations.

\item A prior-guided clustering module is designed to enhance model interpretability by using anatomical prior knowledge and cross-attention to dynamically group individual ROIs into functional communities, thereby effectively modeling the brain's hierarchical interactions.

\item Our method enables identification of disease biomarkers across multiple scales, including ROI-level connections, ROI-to-community affiliations, and inter-community coordination. This capability enhances diagnostic accuracy while ensuring a transparent decision-making process.

\end{itemize}
\begin{figure*}[htb]
  \centering
  \includegraphics[width=\linewidth]{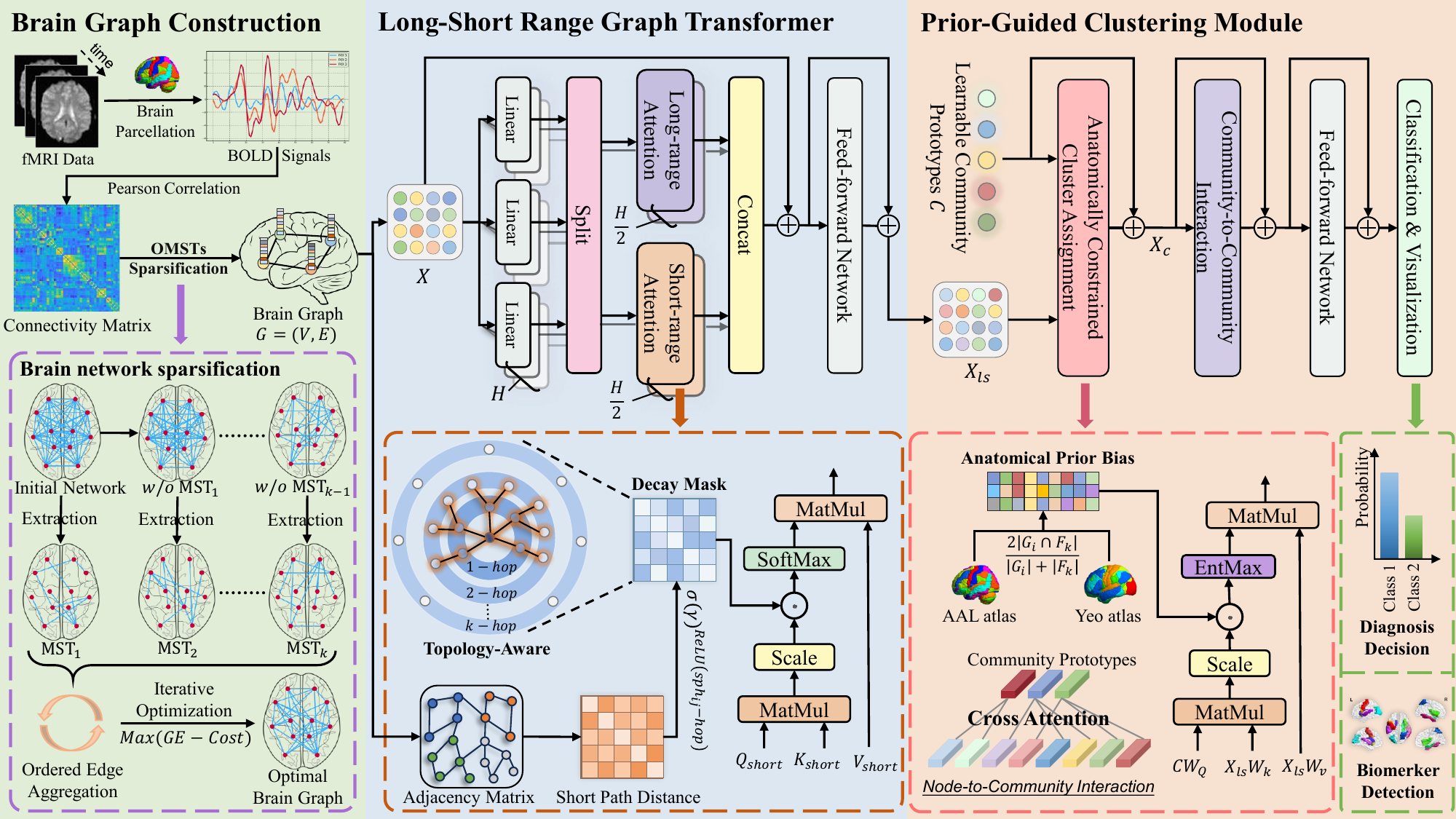}
  \caption{The overall framework of the proposed BrainHGT method. The model first learns multi-scale features of brain regions via a long-short range graph Transformer, and then aggregates these features into biologically plausible functional communities using a prior-guided clustering module for the final classification task.}
  \label{fig:model}
\end{figure*}

\section{Related Work}
\subsection{ROI-Level Brain Network Analysis}
ROI-level studies focus on the interaction patterns among whole-brain ROIs, which is a common paradigm in brain network analysis. Cui et al. \cite{cui2023braingb} introduced BrainGB, a comprehensive benchmark for GNN-based brain network analysis that evaluates the effectiveness of different node features, message-passing mechanisms, and pooling strategies. Zheng et al. \cite{Zheng2024brainIB} developed BrianIB, which uses the information bottleneck principle to identify disease-related subgraphs, improving diagnostic accuracy and providing interpretable biomarkers. However, the local message-passing mechanism of GNNs limits their ability to capture long-range connections. In contrast, Graph Transformer can alleviate this issue by virtue of its global attention mechanism. Peng et al. \cite{peng2024gbt} proposed GBT, which learns representative graph features by applying a low-rank approximation to the attention matrix. Yu et al. \cite{yu2024longrange} proposed ALTER, which adopts random walk to better model long-range dependencies. However, the global attention in these methods often ignores the network's topological structure, leading to an over-globalizing problem and making it difficult to balance the local and global characteristics of brain information processing.

\subsection{Modular-Aware Brain Network Analysis}
Modular-aware studies focus on integrating the community structure of brain function, which corresponds to the sub-functional activities of the brain. This can be broadly classified into two categories. The first category of methods relies on predefined network groups to study functional community interactions. Shehzad et al. developed BrainGT and BrainDGT, which are based on voxel overlap to assign ROIs to functional networks \cite{shehzad2024braingt,shehzad2025dynamic}. Similarly, Bannadabhavi et al. \cite{bannadabhavi2023community} designed Com-BrainTF, which rearranges the functional connectivity matrix based on predefined community labels to learn connection patterns within and between communities. The main limitation of these methods lies in their reliance on predefined groups, which are typically derived from general anatomical or functional divisions. Such groups may fail to align accurately with the true functional communities specific to a given task or disease, thereby constraining the model's flexibility. The second category of methods is dedicated to adaptively learning functional communities directly from data. Li et al. \cite{li2021braingnn} proposed BrainGNN, which utilizes a special graph convolution layer to learn soft membership scores for each brain region, building communities that are relevant to the classification task. Zhu et al. \cite{zhu2022multi} introduced M2CDCA, which utilizes spectral clustering and structural networks to guide community detection in functional networks. Kan et al. \cite{kan2022brain} proposed BRAINNETTF, which draws on the idea of deep embedding clustering, using a self-supervised soft clustering process to group functionally similar brain region nodes. Zhang et al. \cite{zhang2024constructing} proposed HFBN, which features a unique hierarchical node fusion method that adaptively merges fine-grained nodes into coarse-grained nodes layer by layer, thereby constructing a hierarchical brain network. Peng et al. \cite{peng2025biologically} developed BioBGT, which uses the classic Louvain algorithm for unsupervised community detection and introduces a community contrastive learning strategy to optimize the partition by enhancing intra-module integration and inter-module segregation. Although these adaptive methods are more flexible, certain studies lack sufficient guidance from biological priors. This can lead to the learned community structures being unstable or their functional meaning being ambiguous, which poses a challenge to their interpretability in the field of neuroscience.

\section{Method}
The overall framework of the proposed BrainHGT is presented in Figure \ref{fig:model}. First, we construct a sparse and topologically efficient brain graph from fMRI data. Then, a long-short range attention encoder is designed to learn multi-scale node representations that align with brain topology. Finally, an interpretable prior-guided clustering module that aggregates node information into biologically meaningful functional communities to facilitate classification.

\subsection{Brain Graph Construction}
For each subject, the preprocessed fMRI time series from $N$ Regions of Interest (ROIs) are used to compute a Pearson correlation matrix $\mathbf{R} \in \mathbb{R}^{N \times N}$, representing the initial fully-connected functional connectivity network. To preserve essential functional connections while discarding redundant information, we employ the Orthogonal Minimal Spanning Trees algorithm \cite{dimitriadis2017topological} for network sparsification. This constructs a sparse and topologically optimized brain graph that represents the backbone of the brain functional network. The process begins by converting the correlation matrix $\mathbf{R}$ into an inverse weighted graph $\mathbf{D}$, where $d_{ij} = 1 / |r_{ij}|$ (for $r_{ij} \neq 0$), to ensure that stronger functional connections correspond to shorter topological distances. Subsequently, a sequence of orthogonal minimal spanning trees (MSTs), namely $\{\text{MST}_1, \text{MST}_2, \dots, \text{MST}_k\}$, is iteratively extracted from $\mathbf{D}$ using Kruskal's algorithm. In each iteration $i$, the $\text{MST}_i$ with the minimum total edge weight is identified, and its constituent edges are then excluded from the graph for all subsequent iterations to ensure orthogonality. 
Finally, the edges from this sequence of MSTs are aggregated one-by-one in their extraction order, forming a set of candidate networks. The final adjacency matrix $\mathbf{\tilde{A}}$ is the specific candidate network $\mathbf{A}$ that maximizes a global cost-efficiency metric:
\begin{equation}
\mathbf{\tilde{A}} = \arg\max_{\mathbf{A}} \left( \text{GE}(\mathbf{A}) - \text{Cost}(\mathbf{A}) \right).
\end{equation}
Here, Global Efficiency (GE) measures the network's information transfer capability and is defined as the average inverse shortest path length:
\begin{equation}
\text{GE}(\mathbf{A}) = \frac{1}{N(N-1)} \sum_{i \neq j} \frac{1}{L_{ij}},
\end{equation}
where $L_{ij}$ is the shortest path length between nodes $i$ and $j$ in the graph $\mathbf{A}$. Cost represents the wiring cost of the network, defined as the ratio of the total weight of $\mathbf{A}$ normalized by the total weights in the original dense graph:
\begin{equation}
\text{Cost}(\mathbf{A}) = \frac{\sum_{(i,j) \in E_k} |r_{ij}|}{\sum_{i<j} |r_{ij}|}, 
\end{equation} 
where $E_k$ is the set of edges in $\mathbf{A}$. This procedure yields a topologically optimized brain network that preserves core functional pathways while balancing efficiency and cost. As a result, we construct a brain graph $G(V,E)$, where the adjacency matrix is denoted as $\mathbf{\tilde{A}} \in \mathbb{R}^{N \times N}$, and the node features are represented by the correlation vectors from the original matrix $\mathbf{R}$.

\subsection{Long-Short Range Graph Transformer}
The brain’s multi-scale organization features dense local processing within modules and sparse long-range connections for global integration. To learn node representations sensitive to this structure, we introduce the long-short range attention mechanism, designed to overcome the topology-agnostic limitations of the standard Transformer.
LSRA strategically divides the $H$ attention heads into two equal sets: $H_{short} = H/2$ heads dedicated to modeling short-range, topologically-informed interactions, while $H_{long} = H/2$ heads for capturing long-range, global dependencies.

Let $\mathbf{X} \in \mathbb{R}^{N \times d}$ denote the input node feature matrix, where $d$ is the feature dimension. Specifically, $\mathbf{X}$ is derived from a linear mapping of $\mathbf{R}$. The input is projected into query ($\mathbf{Q}$), key ($\mathbf{K}$), and value ($\mathbf{V}$) representations, each of which has a dimension of $\mathbb{R}^{N \times d}$. 
The query, key, and value vectors for the $i$-th node within head $h$ are denoted as $\mathbf{q}_i^{(h)}$, $\mathbf{k}_i^{(h)}$, $\mathbf{v}_i^{(h)}$ respectively, corresponding to the i-th rows of these matrices. 

\subsubsection{Short-Range Attention:}
To focus on interactions between topologically proximal brain regions, the attention scores are modulated by a Topological Decay Mask, derived from the shortest path length (SPL) matrix $\mathbf{S} \in \mathbb{R}^{N \times N}$. For each short-range head $h_s = 1, \dots, H_{short}$, we introduce learnable parameters, a hop threshold $\text{hop}^{(h_s)}$ and a decay factor $\gamma^{(h_s)}$. The raw attention energy $e_{ij}^{(h_s)}$ between node $i$ and node $j$ for head $h_s$ is first computed:
\begin{equation}
    \mathbf{e}_{ij}^{(h_s)} = \frac{(\mathbf{q}_i^{(h_s)})^T \mathbf{k}_j^{(h_s)}}{\sqrt{d_{head}}},
\end{equation}
where $d_{head}$ is the dimension of each attention head. This energy is then modulated by the topological decay:
\begin{equation} 
\mathbf{\hat{e}}_{ij}{}^{(h_s)} = \mathbf{e}_{ij}^{(h_s)} \odot \left(\sigma(\gamma^{(h_s)})\right)^{\text{ReLU}(S_{ij} - \text{hop}^{(h_s)})},
\end{equation}
where $\sigma(\cdot)$ is the Sigmoid function constraining $\gamma^{(h_s)}$ to $(0,1)$, $S_{ij}$ is the SPL between node $i$ and $j$, $\odot$ denotes the Hadamard product. This formulation ensures that attention to nodes beyond the $\text{hop}^{(h_s)}$ distance decays exponentially, focusing the head on local information processing. The final output for head $h_s$ is:
\begin{equation} 
    \mathbf{o}_i^{(h_s)} = \sum_{j=1}^{N} \text{Softmax}_j(\mathbf{\hat{e}}_{ij}{}^{(h_s)}) \mathbf{v}_j^{(h_s)}.
\end{equation}

\subsubsection{Long-Range Attention:}
To capture dependencies between any pair of brain regions irrespective of distance, the long-range heads operate as standard scaled dot-product attention. This facilitates the modeling of global information integration. For each long-range head $h_l = 1, \dots, H_{long}$:
\begin{equation}
    \mathbf{o}_i^{(h_l)} = \sum_{j=1}^{N} \text{Softmax}_j(\frac{(\mathbf{q}_i^{(h_l)})^T \mathbf{k}_j^{(h_l)}}{\sqrt{d_{head}}}) \mathbf{v}_j^{(h_l)}.
\end{equation}

The outputs from all $H$ heads, namely $\{\mathbf{o}_i^{(h_s)}\}_{s=1}^{H_{short}}$ and $\{\mathbf{o}_i^{(h_l)}\}_{l=1}^{H_{long}}$, are concatenated and passed through a final linear projection $\mathbf{W}_O \in \mathbb{R}^{d \times d}$ to produce the output of the LSRA layer:
\begin{equation}
    \mathbf{X_{ls}}(i) = \mathbf{W}_O \left( \text{Concat}(\mathbf{o}_i^{(h_1)}, \dots, \mathbf{o}_i^{(h_H)}) \right).
\end{equation}

This dual-branch design enables the encoder to concurrently process information at distinct spatial granularities, effectively capturing the multi-scale functional architecture inherent in brain networks.

\subsection{Prior-Guided Brain Region Clustering}

To bridge the transition from node-level interactions to community-level insights and simulate brain’s hierarchical information processing, we introduce the prior-guided clustering module. This module groups the $N$ brain regions into $K$ latent functional communities in a data-driven yet biologically-informed manner. To ensure these communities are neuroscientifically meaningful, this process is guided by an explicit anatomical prior.

This process begins by predefining a set of learnable community prototypes $\mathbf{C} \in \mathbb{R}^{K \times d}$, where each vector $\mathbf{c}_k$ represents the initial embedding of a functional community. Inspired by Kan et al \cite{kan2022brain}, the community prototypes are initialized using the Xavier uniform method \cite{glorot2010understanding} and then orthogonalized with the Gram-Schmidt process to enhance the spatial separability of node embeddings from different functional modules. 
To ground the clustering process in established neuroanatomy, we construct a Dice similarity prior matrix, $\mathbf{D}_{\text{prior}} \in \mathbb{R}^{N \times K}$, based on the spatial overlap between input $N$ ROIs and the $K$ canonical large-scale functional networks. 

Let $G_i$ be the set of voxels corresponding to the $i$-th ROI in atlas, and $F_k$ be the set of voxels for the $k$-th functional network. The Dice coefficient between them is:
\begin{equation}
    \mathbf{D}_{\text{prior}}(i,k) = \frac{2 |G_i \cap F_k|}{|G_i| + |F_k|}
\end{equation}
where $|\cdot|$ denotes the number of voxels in a set. Each element $(\mathbf{D}_{\text{prior}})_{ik}$ thus represents a similarity score between ROI $i$ and functional network $k$. This matrix injects valuable biological knowledge, guiding the model to form communities that align with known functional systems.
\begin{table*}[!ht]
\renewcommand{\arraystretch}{1.03}
\centering
\caption{Performance comparison with existing methods on two datasets (Mean ± standard deviation). \textbf{Bold} indicates the best results and \underline{underlining} signifies second outcomes.}
\label{tab:results}
\resizebox{\textwidth}{!}{%
\begin{tabular}{llcccccccc}
\toprule
\multirow{2}{*}{\textbf{Type}} & \multirow{2}{*}{\textbf{Method}} & \multicolumn{4}{c}{\textbf{Dataset: ABIDE}} & \multicolumn{4}{c}{\textbf{Dataset: ADNI}} \\
\cmidrule(lr){3-6} \cmidrule(lr){7-10}
& & \textbf{ACC(\%) $\uparrow$}& \textbf{AUC(\%) $\uparrow$}& \textbf{SEN(\%) $\uparrow$}& \textbf{SPE(\%) $\uparrow$} &  \textbf{ACC(\%) $\uparrow$}& \textbf{AUC(\%) $\uparrow$}& \textbf{SEN(\%) $\uparrow$}& \textbf{SPE(\%) $\uparrow$} \\
\midrule
\multirow{2}{*}{ML Methods} 
& SVM & 60.0$\pm$1.78 & 60.09$\pm$1.88 & 63.15$\pm$4.56 & 57.03$\pm$5.49 & 67.72$\pm$5.43 & 56.76$\pm$3.88 & 85.16$\pm$3.25 & 18.37$\pm$8.75 \\
& Random Forest & 55.25$\pm$2.71 & 55.39$\pm$2.80 & 59.55$\pm$5.74 & 51.23$\pm$7.57 & 64.23$\pm$2.41 & 53.74$\pm$1.87 & 80.59$\pm$5.46 & 16.89$\pm$7.31 \\
\midrule
\multirow{6}{*}{\begin{tabular}{@{}l@{}}Graph Neural \\ Networks\end{tabular}}

& GAT & 60.41$\pm$2.50 & 59.40$\pm$2.48 & 57.96$\pm$11.73 & 59.43$\pm$12.31 & 65.26$\pm$3.66 & 64.22$\pm$3.76 & 80.30$\pm$9.92 & 49.16$\pm$15.96 \\
& BrainGNN & 57.74$\pm$4.40 & 62.10$\pm$3.38 & 39.96$\pm$24.04 & 67.23$\pm$24.29 & 62.14$\pm$3.09 & 65.12$\pm$3.97 & 71.96$\pm$16.05 & 41.24$\pm$14.17 \\
& IBGNN & 59.85$\pm$2.78 & 60.67$\pm$3.27 & 64.56$\pm$13.88 & 53.99$\pm$16.25 & 57.56$\pm$7.49 & 57.12$\pm$4.93 & 64.91$\pm$21.02 & 46.03$\pm$21.21 \\
& BrainGB & 63.12$\pm$3.24 & 66.32$\pm$2.97 & 65.10$\pm$12.01 & 60.40$\pm$8.55 & 67.50$\pm$3.10 & 70.81$\pm$3.70 & 82.07$\pm$9.21 & 43.40$\pm$11.10 \\
& BrainUSL & 62.36$\pm$2.73 & 62.80$\pm$2.40 & 59.21$\pm$10.03 & 66.34$\pm$8.82 & 60.08$\pm$3.32 & 49.64$\pm$3.23 & 79.07$\pm$9.34 & 20.20$\pm$10.08 \\
& BrainIB & 59.47$\pm$2.52 & 62.18$\pm$2.38 &  64.98$\pm$7.86 & 53.26$\pm$6.64 & 65.73$\pm$2.69 & 65.63$\pm$4.15 & 75.37$\pm$4.68 & 47.14$\pm$6.54 \\
\midrule
\multirow{7}{*}{\begin{tabular}{@{}l@{}}Graph \\ Transformer\end{tabular}} 
& VanillaTF & 64.33$\pm$2.70 & 72.97$\pm$2.02 & 62.60$\pm$11.30 & 66.78$\pm$15.80 & 68.46$\pm$2.54 & 72.39$\pm$3.64 & 80.11$\pm$14.26 & \textbf{51.85$\pm$19.61} \\
& BrainNetTF & 67.98$\pm$3.63 & 75.63$\pm$2.74 & \underline{70.31$\pm$15.69} & 61.07$\pm$20.43 & 71.46$\pm$3.75 & 78.02$\pm$2.92 & 83.15$\pm$10.72 & 48.93$\pm$16.91 \\
& GBT & 67.34$\pm$2.37 & 75.67$\pm$3.37 & 65.47$\pm$13.83 & \underline{69.86$\pm$14.29} & 72.20$\pm$2.03 & 77.64$\pm$1.50 & \underline{85.19$\pm$4.90} & 37.50$\pm$9.62 \\
& RGTNet & 63.35$\pm$3.06 & 68.42$\pm$3.50 & 63.83$\pm$10.0 & 62.88$\pm$7.89 & 66.59$\pm$2.63 & 64.96$\pm$4.65 & 76.48$\pm$3.41 & 47.50$\pm$8.60 \\
& Contrasformer & 61.70$\pm$3.18 & 66.75$\pm$3.30 & 54.95$\pm$10.37 & 68.11$\pm$8.20 & 
67.80$\pm$1.90 & 70.60$\pm$2.97 & 70.50$\pm$11.60 & 39.42$\pm$12.85\\
& ALTER & \underline{68.87$\pm$1.89} & \underline{75.80$\pm$2.63} & 69.98$\pm$6.56 & 67.40$\pm$6.81 & \underline{73.29$\pm$2.41} & \underline{79.56$\pm$2.72} & 82.78$\pm$9.37 & 45.20$\pm$21.01 \\
& BioBGT & 63.97$\pm$1.74 & 62.60$\pm$1.77 & 60.91$\pm$6.01 & 65.71$\pm$5.49 & 68.90$\pm$1.84 & 59.48$\pm$4.50 & 78.70$\pm$15.62& 38.21$\pm$14.23 \\
\midrule
Ours & BrainHGT & \textbf{71.33$\pm$1.91} & \textbf{76.65$\pm$1.45} & \textbf{72.40$\pm$3.96} & \textbf{69.93$\pm$2.45} & \textbf{74.27$\pm$1.76}& \textbf{80.84$\pm$1.38}& \textbf{87.22$\pm$5.52}& \underline{49.29$\pm$8.42} \\
\bottomrule
\end{tabular}%
}
\end{table*}

\subsubsection{Anatomically Constrained Cluster Assignment:}
The core of the module is a cross-attention mechanism that computes the soft assignment of nodes to communities. Here, the community prototypes $\mathbf{C}$ act as queries, while the node features $\mathbf{X}_{ls} \in \mathbb{R}^{N \times d}$ from the LSRA encoder serve as keys and values. The attention energy between community $k$ and ROI $i$ is modulated by the Dice prior.
\begin{equation}
    \mathbf{\hat{e}}_{ki} = \frac{(\mathbf{c}_k \mathbf{W}_Q)((\mathbf{x}_{ls})_i \mathbf{W}_K)^T}{\sqrt{d_{head}}} \odot (\mathbf{D}_{\text{prior}})_{ik}
\end{equation}
where $\mathbf{W}_Q, \mathbf{W}_K \in \mathbb{R}^{d \times d}$ are learnable projection matrices and $d_{head}$ is the dimension of each attention head.
This step ensures that a high feature similarity ($\mathbf{\hat{e}}_{ki}$) is amplified if it corresponds to a high spatial overlap ($(\mathbf{D}_{\text{prior}})_{ik}$), and suppressed otherwise.
To obtain a sparse and more definitive assignment distribution, we apply the $\text{Entmax}$ function transformation to the biased scores. The resulting attention weights form the soft assignment matrix $\mathbf{P} \in \mathbb{R}^{N \times K}$:
\begin{equation}
    \mathbf{P}[i,:] = \text{Entmax}((\mathbf{\mathbf{\hat{e}}}[:,i])^T)
\end{equation}
The updated representation for each community, $\mathbf{\hat{c}}_k$, is then computed by taking the weighted sum of node value vectors: 
$\mathbf{\hat{c}}_k = \sum_{i=1}^{N} \mathbf{P}_{ik} ((\mathbf{x}_{ls})_i \mathbf{W}_V)$, where  $\mathbf{W}_V\in\mathbb{R}^{d \times d}$ is learnable projection matrices.

\subsubsection{Refinement of Community Interaction.}
To further model community-level coordination pattern, these aggregated community representations $\mathbf{X}_c=\{\mathbf{\hat{c}}_k\}_{k=1}^K\in\mathbb{R}^{K \times d}$ are then refined through a self-attention layer followed by a feed-forward network. This crucial step models inter-community dynamics, allowing the framework to capture complex relationships between the identified functional networks. These hierarchical-aware community representations are then aggregated via mean pooling and passed through a 256-32-2 multilayer perceptron for final classification.

By utilizing a hierarchical architecture that first learns multi-scale node representations and then aggregates them into meaningful functional communities. This multi-stage process produces a final representation that is optimized for classification while remaining highly interpretable.

\section{Experiments and Results}
\subsection{Materials}
We evaluate our proposed method on two commonly used public fMRI datasets.

\subsubsection{ABIDE Dataset.} The ABIDE dataset is a database for ASD research, comprising data from 17 international sites \cite{craddock2013neuro,liu2024randomizing}. In total, 1009 subjects are utilized, including 516 individuals diagnosed with ASD and 493 normal controls (NC). The brain is parcellated into 200 ROIs using the Craddock 200 (CC200) atlas \cite{craddock2012whole}. 

\subsubsection{ADNI Dataset.} The ADNI dataset is a longitudinal data study focused on discovering early biomarkers for AD \cite{petersen2010alzheimer,zhang2024asynchronous}. A cohort of 410 subjects is utilized, including 263 patients with mild cognitive impairment (MCI) and 147 normal controls, this carefully matches for both age and sex ratio. The Automated Anatomical Labeling (AAL) atlas \cite{tzourio2002automated} is employed to define 90 ROIs for the brain of each subject.

The preprocessing of fMRI data follows the standardized pipeline \cite{shehzad2015preprocessed, yan2016dpabi, esteban2019fmriprep}. The anatomical prior serves as a key component of the Prior-Guided Brain Region Clustering module. Here, we utilize the Yeo 7-network atlas \cite{yeo2011organization} to cluster the ROIs into eight sub-networks as a prior, they are Default Mode Network (DMN), Frontoparietal Network (FPN), Limbic Network (LN), Ventral Attention Network (VAN), Dorsal Attention Network (DAN), Somatomotor Network (SMN), Visual Network (VN), and Cerebellum and Subcortical Structures (CB \& SC).

\subsection{Experimental Settings}
All experiments were conducted on a GeForce GTX 3080 Ti GPU, with implementation in PyTorch v1.12.1. For training, we employed the Adam optimizer with an initial learning rate of 1e-4 and a weight decay of 1e-4. The model was trained for 100 epochs with a batch size of 32, and the checkpoint that achieved the highest Area Under the Receiver Operating Characteristic Curve (AUC) on the validation set was selected for the final performance evaluation on the test set. More details of hyperparameters and model complexity refer to Appendix \textbf{C} and Appendix \textbf{F}, respectively.

For each dataset, we followed a consistent evaluation protocol by randomly splitting the data into 70\% for training, 10\% for validation, and 20\% for testing. This process was repeated 10 times, with the results averaged to ensure robustness and reliability.

\subsection{Comparison with Existing Methods}
We compare our BrainHGT with three types of methods. 1) typical machine learning (ML) methods, including SVM (with a linear kernel) and Random Forest; 2) GNN-based methods, including GAT \cite{velickovic2017graph}, BrainGNN \cite{li2021braingnn}, IBGNN \cite{cui2022interpretable}, BrainGB \cite{cui2023braingb}, BrainUSL \cite{zhang2023brainusl} and BrainIB \cite{Zheng2024brainIB}. 3) Graph Transformer-based methods, including vanillaTF \cite{kan2022brain}, BrainNetTF \cite{kan2022brain}, GBT \cite{peng2024gbt}, ALTER \cite{yu2024longrange}, RGTNet \cite{wang2024residual}, Contrasformer \cite{xu2024contrasformer} and BioBGT \cite{peng2025biologically}. For these models, we used the official open-source code and its default configuration.

Table \ref{tab:results} summarizes the experimental results on the two datasets,  
using Accuracy (ACC), AUC, Sensitivity (SEN), and Specificity (SPE) as the primary metrics.
As can be seen, our method achieves the best performance across most metrics on both datasets, which strongly validates that simulating the brain's hierarchical functional structure is a more effective strategy for learning discriminative features for disease classification. 

In addition, deep learning methods that model brain connectome topology consistently outperform traditional methods, highlighting the importance of learning features directly from the network structure. And Graph Transformer-based methods generally surpass those GNN-based methods by overcoming the latter's reliance on local message passing to better capture global dependencies. 

\begin{table}[t]
    \centering
    \renewcommand{\arraystretch}{1.3}
    \setlength{\tabcolsep}{3pt}
    \caption{Ablation Studies of different components on two disease classification tasks. \textbf{Bold} indicates the best results.}
    \resizebox{1.0\linewidth}{!}{
    \begin{tabular}{c|c|cccc}
        \toprule
         \textbf{Datasets} & \textbf{Component} & \textbf{ACC(\%) $\uparrow$}& \textbf{AUC(\%) $\uparrow$}& \textbf{SEN(\%) $\uparrow$}& \textbf{SPE(\%) $\uparrow$}\\
        \midrule
        \multirow{5}{*}{\textbf{ABIDE}} 
        & w/o LSRA &69.99$\pm$3.06  &75.29$\pm$3.17 & 71.64$\pm$3.95&68.0$\pm$5.01\\
        & w/o $\mathbf{D}_{\text{prior}}$ &70.20$\pm$1.70  & 75.78$\pm$1.58 & 70.99$\pm$4.66& 69.16$\pm$5.06\\
        & w/o Entmax &70.0$\pm$2.16  & 75.73$\pm$1.93 & 71.95$\pm$4.70& 67.47$\pm$4.83\\
        & w/o Clustering &65.02$\pm$1.69  & 69.85$\pm$1.83 & 70.27$\pm$11.6& 58.45$\pm$11.6\\
        & BrainHGT  &\textbf{71.33$\pm$1.91} & \textbf{76.65$\pm$1.45} & \textbf{72.40$\pm$3.96} & \textbf{69.93$\pm$2.45}\\
        \midrule
        \multirow{5}{*}{\textbf{ADNI}} 
        & w/o LSRA &72.80$\pm$1.89  &79.12$\pm$1.85 & 81.74$\pm$3.71 &47.86$\pm$3.98\\
        & w/o $\mathbf{D}_{\text{prior}}$ &73.17$\pm$2.44  &80.21$\pm$2.27 & 83.51$\pm$5.33 &53.21$\pm$9.90\\
        & w/o Entmax &72.20$\pm$2.30 & 79.65$\pm$2.37 & 80.81$\pm$3.49& 47.86$\pm$5.80\\
        & w/o Clustering &69.88$\pm$3.41  & 73.55$\pm$4.23 & 76.48$\pm$4.27& \textbf{57.14$\pm$10.71}\\
        & BrainHGT  &\textbf{74.27$\pm$1.76}& \textbf{80.84$\pm$1.38}& \textbf{87.22$\pm$5.52}& 49.29$\pm$8.42\\
        \bottomrule
    \end{tabular}
    }
    \label{tab:ablation}
\end{table}

\subsection{Ablation Studies}
We conduct ablation studies on two datasets to validate the effectiveness of the designed components in BrainHGT. As shown in Table \ref{tab:ablation}, replacing our long-short range graph Transformer with a standard Transformer (w/o LSRA) degrades performance, demonstrating the crucial role of LSRA in balancing local and global information processing. Removing the anatomical constraint (w/o $\mathbf{D}_{\text{prior}}$) or replacing Entmax with Softmax for generating soft cluster assignments (w/o Entmax) also leads to a decline in performance, which confirms the importance of biological guidance and sparse assignment strategies. Notably, the most significant performance drop occurs when the entire hierarchical clustering module is removed (w/o Clustering), strongly indicating that modeling the brain's hierarchical functional structure is central to our model's superior performance.

\begin{figure}[t]
\centering
\includegraphics[width=1\columnwidth]{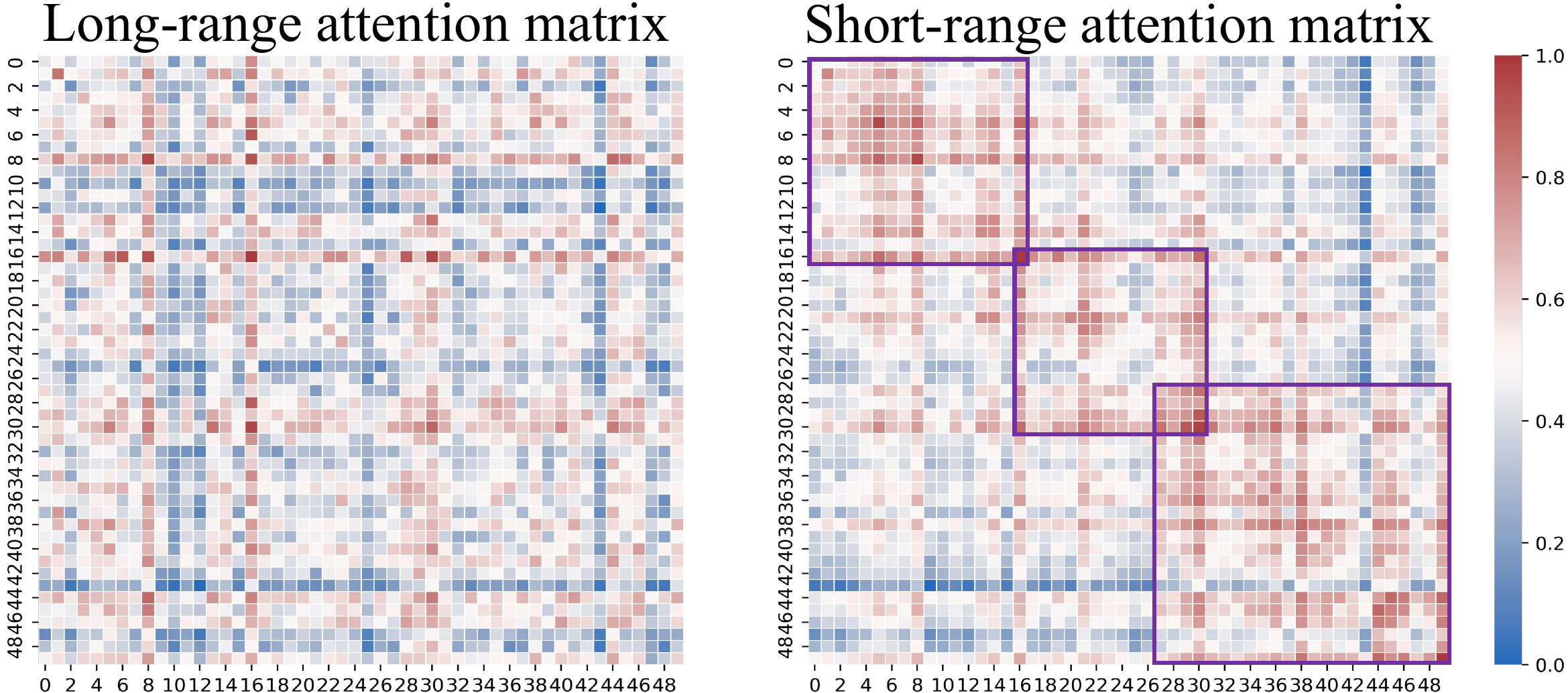} 
\caption{Visualizing attention score from the long-short range attention module.}
\label{fig:lsra}
\end{figure}

\begin{figure*}[htb]
  \centering
  \includegraphics[width=0.962 \linewidth]{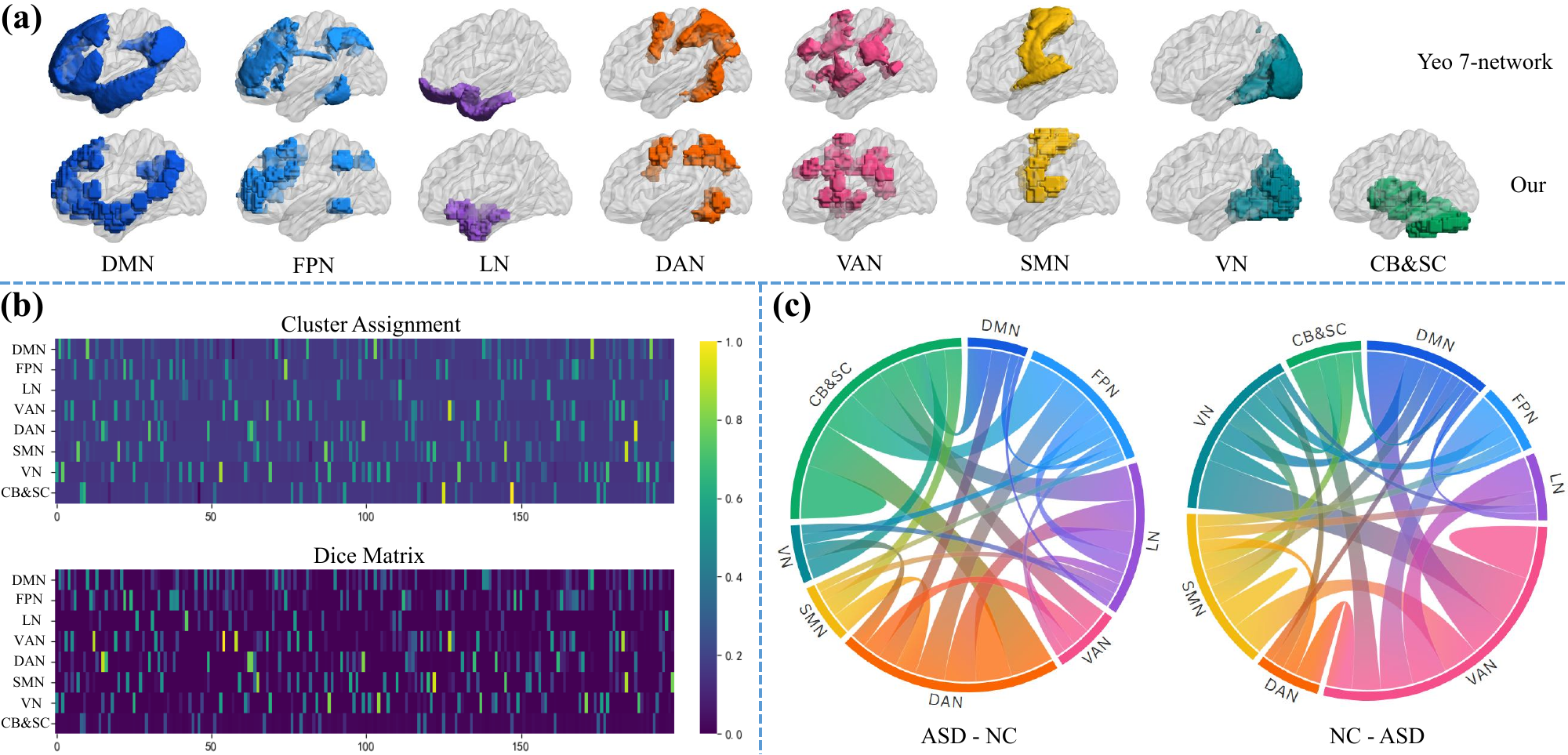}
  \caption{(a) Comparison of our learned functional communities with the standard Yeo 7-network. (b) Visualizing soft cluster assignment matrix and Dice prior matrix. (c) Differential community interactions highlight hyper-connectivity (left) and hypo-connectivity (right) in ASD.}
  \label{fig:cluster_bio}
\end{figure*}

\subsection{Interpretability and Visualization}
To validate the interpretability of our method, we analyze correctly classified samples from test set to identify potential biomarkers and visualize its internal decision-making process. Without loss of generality, we use the ABIDE dataset for illustration.

\subsubsection{Analysis of ROI-level Interaction Patterns.}
Figure \ref{fig:lsra} visualizes the long-range and short-range attention from the LSRA module. The long-range matrix shows a diffuse pattern to capture global dependencies, while the short-range matrix has a clear block or banded structure focused on local interactions. This indicates that the short-range attention is constrained by the topological decay mask, whereas the long-range attention operates free from spatial restrictions. This observation verifies the necessity of LSRA in segregating information processing across different spatial scales.
\subsubsection{Visualization of Learned Functional Communities.}
To evaluate the biological plausibility of the learned functional communities, we first extracted the soft assignment matrix from the prior-guided clustering module, then assigned each ROI to its highest-probability community for visualization (Specific groupings refer to Appendix \textbf{D}). As shown in Figure \ref{fig:cluster_bio} (a), the resulting functional communities demonstrate high spatial correspondence with the standard Yeo-7 atlas, intuitively confirming their biological plausibility. 
However, this correspondence is not a simple replication of the anatomical prior but performs a data-driven refinement.
As shown in Figure \ref{fig:cluster_bio} (b), the learned clustering assignment is similar to the Dice prior matrix while incorporating disease-relevant characteristics. The model transforms the rigid anatomical constraints into a soft assignment relationship where each ROI can contribute to multiple communities to varying degrees. This flexibility is key to capturing the brain's functional complexity and achieving an adaptive partitioning of the connectome.

\begin{table}[t]
    \centering
    \renewcommand{\arraystretch}{1.3}
    \setlength{\tabcolsep}{3pt}
    \caption{Model performance comparison using different atlas as the anatomical prior. \textbf{Bold} indicates the best results.}
    \resizebox{1.0\linewidth}{!}{
    \begin{tabular}{c|c|cccc}
        \toprule
         \textbf{Datasets} & \textbf{Atlas} & \textbf{ACC(\%) $\uparrow$}& \textbf{AUC(\%) $\uparrow$}& \textbf{SEN(\%) $\uparrow$}& \textbf{SPE(\%) $\uparrow$}\\
        \midrule
        \multirow{2}{*}{\textbf{ABIDE}} 
        & Yeo 7-network &\textbf{71.33$\pm$1.91} & \textbf{76.65$\pm$1.45} & \textbf{72.40$\pm$3.96} & \textbf{69.93$\pm$2.45}\\ 
        & Yeo 17-network &70.10$\pm$1.38  & 75.54$\pm$1.95 & 71.12$\pm$2.57& 68.73$\pm$3.97\\
        \midrule
        \multirow{2}{*}{\textbf{ADNI}} 
        & Yeo 7-network &74.27$\pm$1.76& \textbf{80.84$\pm$1.38} & 87.22$\pm$5.52& \textbf{49.29$\pm$8.42}\\
        & Yeo 17-network &\textbf{74.76$\pm$1.54}  &80.63$\pm$2.51 & \textbf{88.70$\pm$4.18} &47.86$\pm$5.35\\
        \bottomrule
    \end{tabular}
    }
    \label{tab:generalization}
\end{table}
\subsubsection{Identifying Disease-Relevant Communities.} 
To identify community-level biomarkers, we analyzed the differential interaction patterns between the ASD and NC groups by subtracting their average inter-community attention weights. As shown in Figure \ref{fig:cluster_bio} (c), our model discovered patterns of aberrant connectivity that are highly consistent with the neuroscience literature. It revealed significant hyper-connectivity between the cerebellum and cortical association networks like the LN and DAN, consistent with extensive findings of abnormal cerebellar-cortical circuits in ASD \cite{d2015cerebro,cerliani2015increased}. Concurrently, the model identified prominent hypo-connectivity primarily on the VAN, with decreased interaction with VN and DMN, aligning with previously reported findings \cite{duan2017resting,anteraper2020functional}. This automated discovery of clinically-relevant biomarkers highlights our model's potential for understanding complex neurological disorders.

\subsection{Influence of Anatomical Prior Atlases}
To evaluate the model's robustness to the choice of anatomical prior, we evaluated its performance using both the Yeo 7-network and the more fine-grained Yeo 17-network atlases \cite{yeo2011organization}. As shown in Table \ref{tab:generalization}, both atlases yield improved performance, confirming the model's robustness to atlas selection. In addition, the coarser 7-network atlas yields superior performance on the ABIDE dataset, whereas results are largely comparable on the ADNI dataset, this suggests the optimal prior granularity may be task-dependent.

\section{Conclusion}
In this work, we propose a novel hierarchical Graph Transformer framework that enhances brain network analysis by simulating the brain's natural information processing from local regions to global communities. The long-short range attention module mitigates the over-globalizing problem by processing dense local interactions and sparse global connections in parallel, yielding richer node representations. The prior-guided clustering module leverages anatomical prior knowledge to guide the dynamic partitioning of functional communities while modeling hierarchical interaction, significantly improving the model's biological plausibility and interpretability. Our method not only achieves superior performance but also reliably identifies key biomarkers associated with neurological disorders.
\appendix
\section{Acknowledgments}
This work was supported in part by the Natural Science Research Project of Anhui Educational Committee (2023AH050081), the National Natural Science Foundation of China (No. 62476004), the Collaborative Innovation Project of Key Laboratory of Philosophy and Social Sciences in Anhui Province (HZ2302), the Scientific Research Project of Colleges and Universities in Anhui Province (2024AH040115) and the Anhui Province Science and Technology Innovation and Tackling Key Problems Project (202423k09020041).

\bigskip

\bibliography{aaai2026}

\appendix

\section{Appendix}

\subsection{A. \quad Sensitivity Analysis of Hop Value in Short-Range Attention}
\begin{figure}[h]
  \centering
  \includegraphics[width=1\linewidth]{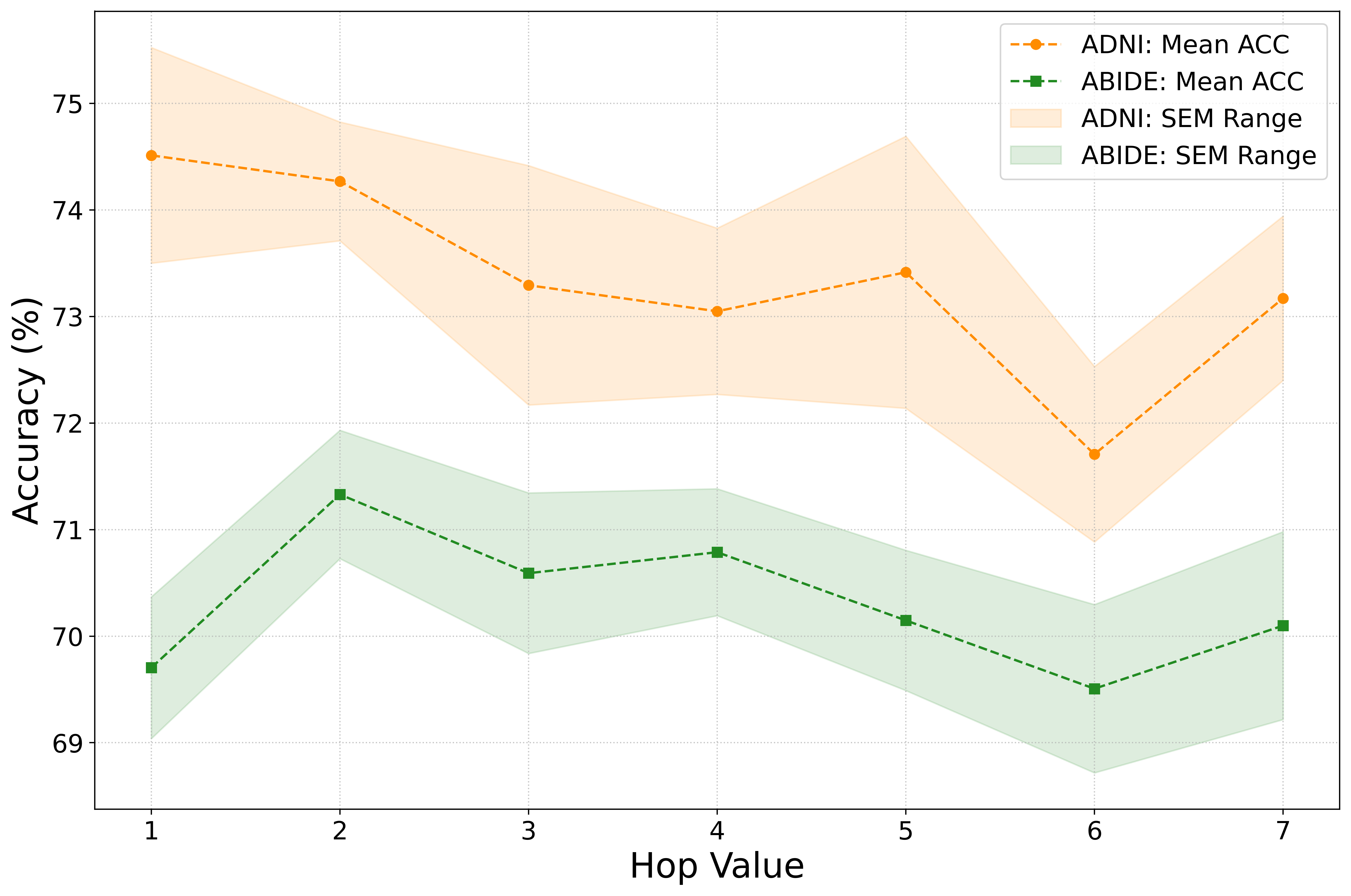}
  \caption{The performance of different hop value on two datasets.}
  \label{fig:hop}
\end{figure}

Figure \ref{fig:hop} illustrates the model's performance sensitivity to the hyperparameter hop in the short-range attention mechanism. We evaluated the model's performance across a range of discrete hop values. The results show that performance peaked on both datasets when the hop value was set to 2.
This outcome aligns with expectations. A hop value that is too small restricts the model's local receptive field, preventing it from capturing sufficient information from functionally relevant neighbors. Conversely, a hop value that is too large can introduce noise from less relevant and distant nodes, thereby diluting the attention weights and degrading performance. Therefore, a hop value of 2 was selected as the optimal parameter, as it strikes the best balance between capturing necessary local interaction and avoiding noise.

\subsection{B. \quad Comparison of Brain Network Sparsification Methods}
\begin{figure}[h]
  \centering
  \includegraphics[width=1.\linewidth]{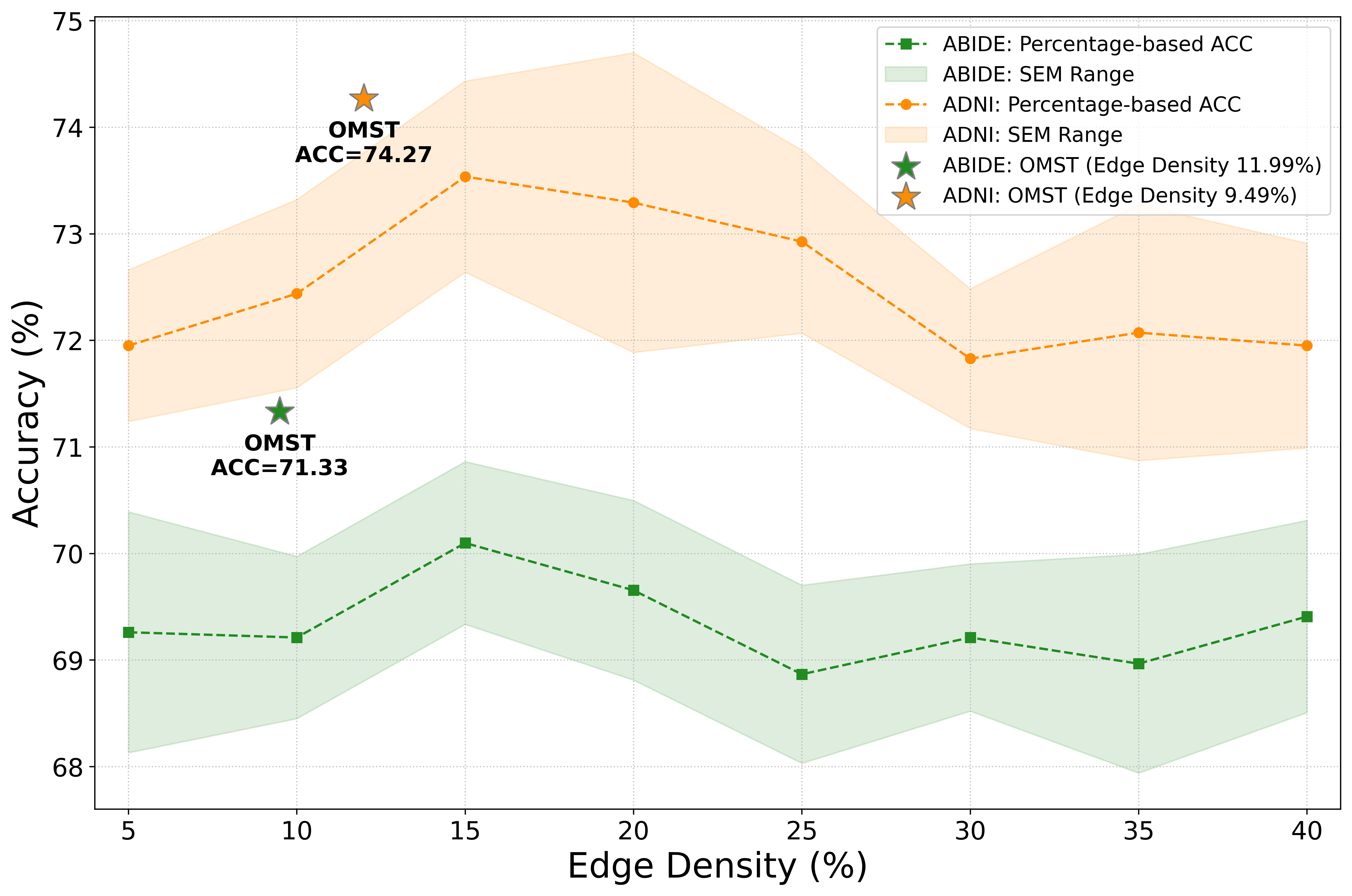}
  \caption{The performance of different network sparsification methods on two datasets.}
  \label{fig:sparsification}
\end{figure}

To validate the effectiveness of Orthogonal Minimal Spanning Trees (OMSTs) method, Figure \ref{fig:sparsification} compares its performance with a conventional percentage-based thresholding sparsification approach, which retains only the strongest connections. The average network densities for graphs constructed by OMST were 9.49\% on the ADNI dataset and 11.99\% on the ABIDE dataset. In contrast, the threshold-based method achieved its peak performance at a 15\% edge density on both datasets.
As indicated by the stars in the figure, the OMST method consistently outperformed the percentage-based approach across the different network densities on both datasets. This result demonstrates that OMSTs algorithm more effectively preserves critical functional connections, yielding brain networks with higher topological efficiency and thereby improving the accuracy of the downstream classification task.

\subsection{C. \quad Implementation Details}
The detailed hyperparameter settings for training BrainHGT on two datasets are summarized in Table \ref{tab:hyperparameters}

\begin{table}[h]
\centering
\caption{Hyperparameters for training BrainHGT on two datasets.} 
\label{tab:hyperparameters}
    \begin{tabular}{l|cc}
    \toprule
    \textbf{Hyperparameter} & \textbf{ABIDE} & \textbf{ADNI}  \\
    \midrule
    \#Layers & 1 & 1  \\
    \#Attention heads & 8 & 8  \\
    Hop value         & 2 & 2  \\
    \#Cluster center & 8 & 8  \\
    The $\alpha$ value in Entmax & 1.5 & 1.5\\
    Hidden dimensions & 256 & 256  \\
    FFN hidden dimensions & 1024 & 1024  \\
    Dropout rate& 0.1 & 0.2  \\
    Readout method & mean & mean\\
    Batch size & 32 & 32 \\
    \#Epochs & 100 & 100  \\
    Learning rate & 1e-4 & 1e-4  \\
    Weight decay & 1e-4 & 1e-4  \\
    \bottomrule
    \end{tabular}
\end{table}

\subsection{D. \quad Functional Module and Brain Region Assignments}
Table \ref{tab:roi_assignments} provides a detailed mapping of the eight functional communities (e.g., DMN, FPN) learned by our model on the ABIDE dataset. The table details the composition of each functional community by listing its constituent ROIs, which are identified by their names and indices from the CC200 atlas. 

\begin{table*}[t]
    \centering
    \caption{Functional Module and Brain Region Assignments}
    \label{tab:roi_assignments}
    
    \begin{tabularx}{\textwidth}{@{} >{\centering\arraybackslash}m{0.16\textwidth} | >{\centering\arraybackslash}X @{}}
    \toprule
        \multirow{1}{*}{\textbf{Functional Module}} & \multirow{1}{*}{\textbf{Brain Region (Name \& Index)}} \\
    \midrule
        \multirow{6}{*}{\textbf{DMN}} & lANG.1 (2), lACC.1 (5), lMTG.1 (11), rANG.2 (14), rACC.1 (22), rPCC (46), rMTG.2 (49), lACC.2 (51), rFOG.1 (53), rPRECU.3 (58), lMFG.2 (61), lMTG.2 (72), lFOG.2 (74), lANG.3 (82), rMEDFG.2 (91), lITG.2 (101), lFP.2 (104), rMFG.4 (106), rMTG.4 (107), rFP.3 (109), lMTG.3 (117), lSTG.2 (129), lSFG.3 (133), rFP.5 (139), rMTG.5 (140), lIFG.1 (141), rIFG.2 (144), rSTG.3 (153), rANG.3 (166), lIFG.2 (167), lSFG.4 (173), lPRECU.3 (174), rTP.3 (181), rSFG.2 (186), rSFG.3 (187), lSFG.5 (191), rSFG.4 (193), lINS.5 (196) \\ 
    \midrule
        \multirow{4}{*}{\textbf{FPN}} & rANG.1 (7), rFP.1 (12), lMFG.1 (23), rMFG.1 (25), lpreC.1 (34), rMFG.2 (38), rMTG.1 (39), rACC.2 (40), lFP.1 (42), lANG.2 (56), rFP.2 (75), lSFG.2 (95), lITG.1 (99), rMFG.5 (113), rpreC.4 (115), rFP.4 (124), rMFG.6 (127), lMEDFG.1 (149), lMFG.4 (151), rIFG.3 (164), lIFG.3 (169), lFP.3 (183) \\ 
    \midrule
        \multirow{2}{*}{\textbf{LN}} & lPARAH.1 (27), rTP.1 (32), lTP.1 (43), lFOG.1 (57), rFOG.2 (71), lTP.2 (78), rFUS.2 (87), rTP.2 (110), lFOG.3 (112), lPARAH.3 (145), rPARAH.2 (155), rACC.4 (160), rFUS.5 (198) \\ 
    \midrule
        \multirow{4}{*}{\textbf{VAN}} & lINS.1 (4), rPRECU.2 (6), rMEDFG.1 (13), lINS.2 (20), rCING (29), lSUPRAM (33), rACC.3 (55), rINS.2 (59), rMTG.3 (69), lPCC (76), lACC.3 (79), rSTG.2 (83), lpostC.5 (116), rIFG.1 (119), lMFG.3 (125), rIPG (128), rINS.3 (137), rSFG.1 (161), lpreC.4 (165), rMFG.7 (168), rpostC.4 (180), lSMA (182), lINS.4 (184) \\
    \midrule
        \multirow{3}{*}{\textbf{DAN}} & lIOG.2 (16), rpreC.1 (17), rSPL.1 (31), lSFG.1 (50), lFUS.1 (63), rMFG.3 (64), lpreC.2 (73), rSUPRAM (93), rITG (100), lSOG.2 (114), rSPG (132), lSOG.3 (136), rIOG.4 (143), lSPL.1 (156), rPRECU.4 (163), lSPL.2 (171), lSPL.3 (188), lPRECU.4 (197) \\ 
    \midrule
        \multirow{3}{*}{\textbf{SMN}} & lpostC.1 (8), lpostC.2 (21), lSTG.1 (24), rpostC.1 (28), rINS.1 (35), rpostC.2 (60), rSPL.2 (65), rSTG.1 (66), lpostC.3 (88), lpreC.3 (90), lpostC.4 (96), rpreC.2 (98), rpreC.3 (111), lINS.3 (121), rparaC (123), rpreC.5 (134), lSTG.3 (146), lpostC.6 (154), rpostC.3 (157), rTTG (185), lTTG (200) \\
    \midrule
        \multirow{5}{*}{\textbf{VN}} & lIOG.1 (1), rPRECU.1 (3), lPRECU.1 (19), rIOG.1 (26), rOP.1 (44), rPARAH.1 (48), lCUN (54), rFUS.1 (62), lLING.1 (70), rCUN (81), rIOG.2 (85), rLING.1 (89), lSOG.1 (97), rIOG.3 (102), rLING.2 (105), lLING.2 (108), lPARAH.2 (122), lIOG.3 (131), rFUS.3 (138), rMOG (142), lPRECU.2 (147), lIOG.4 (150), lOP (158), lFUS.2 (159), rSOG (170), rFUS.4 (172), rIOG.5 (175), lLING.3 (177), lLING.4 (179), lFUS.3 (189), rLING.3 (195) \\
    \midrule
        \multirow{6}{*}{\textbf{CB\&SC}} & lCEREB.1 (9), rCEREB.1 (10), rPUT.1 (15), rTHAL.1 (18), rBSTEM.1 (30), lCEREB.2 (36), lTHAL.1 (37), rCEREB.2 (41), lTHAL.2 (45), lCAUD.1 (47), lBSTEM.1 (52), lPUT (67), rMIDB.1 (68), rCEREB.3 (77), lCEREB.3 (80), lCAUD.2 (84), rBSTEM.2 (86), lAMY (92), rCAUD.1 (94), lCEREB.4 (103), rCEREB.4 (118), lCEREB.5 (120), rCEREB.5 (126), rMIDB.2 (130), rCAUD.2 (135), rTHAL.2 (148), rCEREB.6 (152), lCEREB.6 (162), rCEREB.7 (176), rPUT.2 (178), rBSTEM.3 (190), lCEREB.7 (192), rMIDB.3 (194), lBSTEM.2 (199) \\
    \bottomrule
    \end{tabularx}
\end{table*}
\begin{figure}[tp]
  \centering
  \includegraphics[width=.845 \linewidth]{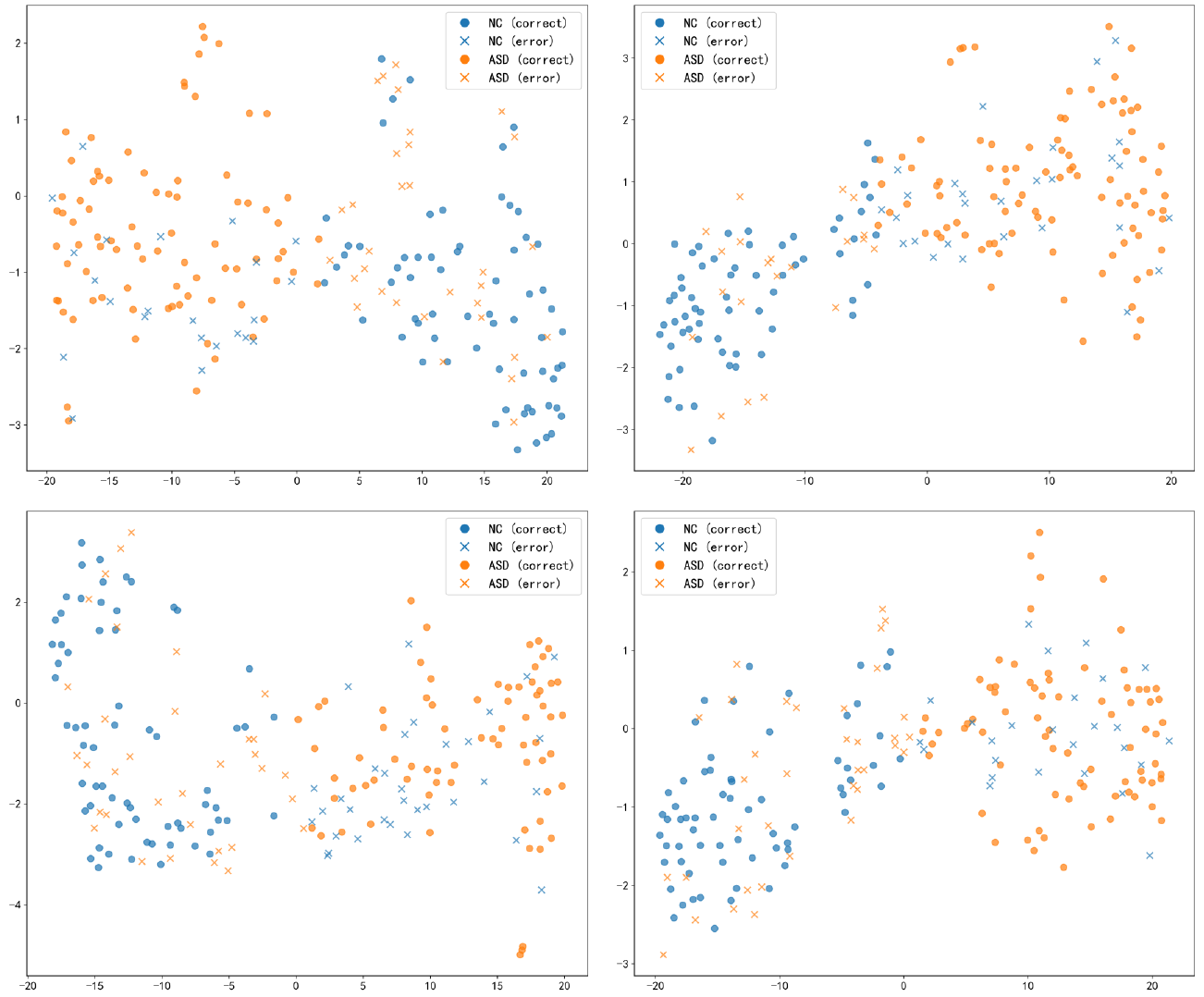}
  \caption{The t-SNE visualization of our proposed method on ABIDE test set.}
  \label{fig:tsen}
\end{figure}
\subsection{E. \quad Visualization of Feature Embeddings}
Figure \ref{fig:tsen} illustrates the t-SNE visualization of the feature embeddings learned from the ABIDE test set. Each point represents a subject, with different colors and shapes distinguishing between the NC and ASD groups, as well as indicating correct or incorrect classifications. The clear separation between the clusters for the two groups demonstrates that the features learned by our model are highly discriminative.

\subsection{F. \quad Analysis of Model Complexity and Runtime}
As shown in Table \ref{tab:parameters}, our BrainHGT model has a higher parameter count compared to other methods. We consider this increased complexity a necessary trade-off for the model's enhanced performance and interpretability, which are derived from its hierarchical architecture. Despite its larger size, the model effectively avoids overfitting due to strong inductive biases. Specifically, the LSRA mechanism and the integration of anatomical priors act as powerful regularizers, guiding the model to learn neuroscientifically plausible features.

Furthermore, BrainHGT is computationally efficient. Its time complexity is determined by its attention mechanisms. The LSRA module has a complexity of $O(N^2 \cdot d)$, while the clustering module's is $O(d \cdot (K \cdot N + K^2))$, where N is the number of nodes and K is the number of communities. Since $N \gg K$, the overall complexity is dominated by the LSRA module, resulting in $O(N^2 \cdot d)$. This is consistent with standard Transformer architectures and demonstrates a favorable balance between performance, interpretability, and computational cost.

\begin{table}[h]
\centering
\caption{Comparison of the number of parameters and running times for different models on the ABIDE and ADNI datasets.}
\label{tab:parameters}
\setlength{\tabcolsep}{4pt}
\resizebox{1.0\linewidth}{!}{
\begin{tabular}{l|rcrc}
\toprule
 \multirow{2}{*}{\textbf{Method}} & \multicolumn{2}{c}{\textbf{Dataset: ABIDE}} & \multicolumn{2}{c}{\textbf{Dataset: ADNI}} \\
\cmidrule(lr){2-3} \cmidrule(lr){4-5}
& \textbf{\#Para}& \textbf{Time(min)}& \textbf{\#Para}& \textbf{Time(min)} \\
\midrule

VanillaTF & 1.56M & 10.60 & 1.36M & 3.60 \\
RGTNet & 0.15M & 19.78 & 0.04M & 6.78 \\
Contrasformer & 1.35M & 139.0 & 0.93M & 36.34 \\
BrainNetTF & 4.0M & 12.92 & 2.57M & 4.27 \\
GBT & 4.0M & 157.87 & 2.57M & 30.78 \\
ALTER & 4.64M & 35.47 & 3.0M & 10.45 \\
BioBGT & 0.69M & 233.67 & 0.46M & 87.55 \\
BrainHGT & 5.17M & 16.84 & 3.31M & 4.13 \\
\bottomrule
\end{tabular}
}
\end{table}

\end{document}